%% file: main.tex
\documentclass[preprint, 3p, times, authoryear, nonatbib]{elsarticle}

\usepackage[utf8]{inputenc}
\usepackage[T1]{fontenc}
\usepackage{amsmath, amssymb}
\usepackage{graphicx}
\usepackage{booktabs}
\usepackage{siunitx}
\usepackage{url}
\usepackage{hyperref}
\usepackage{xcolor}
\usepackage{subcaption}

\graphicspath{{figures/}}

\usepackage[backend=bibtex, style=authoryear-comp, defernumbers=true, maxcitenames=2, uniquelist=false]{biblatex}
\DeclareFieldFormat{labelnumberwidth}{#1\adddot}
\defbibenvironment{bibliography}
  {\list
     {\printtext[labelnumberwidth]{\printfield{labelnumber}}}
     {\setlength{\labelwidth}{\labelnumberwidth}%
      \setlength{\leftmargin}{\labelwidth}%
      \setlength{\labelsep}{\biblabelsep}%
      \addtolength{\leftmargin}{\labelsep}%
      \setlength{\itemsep}{\bibitemsep}%
      \setlength{\parsep}{\bibparsep}}%
      }
  {\endlist}
  {\item}

\renewbibmacro*{cite:labeldate+extradate}{%
  \iffieldundef{labelyear}
    {}
    {\printtext[bibhyperref]{\printlabeldateextra}}%
  \setunit{\addspace}%
  \printtext[bibhyperref]{\mkbibbrackets{\printfield{labelnumber}}}}

\journal{Accident Analysis and Prevention}

\begin{document}

\begin{frontmatter}

\title{The Harsh Truth: Segment-Level Analysis of Harsh Driving Events in Milan Using Large-Scale Telematics, Street Networks, and Google Street View}

\author[mit]{Andrea La Grotteria}
\ead{lagro@mit.edu}

\author[mit,CNR]{Paolo Santi}

\author[mit,AMS]{Titus Venverloo}

\author[mit]{Umberto Fugiglando}

\author[mit,polimi]{Carlo Ratti}

\affiliation[mit]{%
  organization={Senseable City Lab, Massachusetts Institute of Technology},
  addressline={77 Massachusetts Avenue},
  city={Cambridge},
  postcode={02139},
  state={MA},
  country={USA}}

  \affiliation[polimi]{%
  organization={Politecnico di Milano, Department ABC},
  addressline={Via Giuseppe Ponzio 31},
  city={Milano},
  postcode={20133},
  country={Italy}}

 \affiliation[CNR]{%
  organization={Consiglio Nazionale delle Ricerche},
  addressline={Via G. Moruzzi 1},
  city={Pisa},
  postcode={56124},
  country={Italy}}

  \affiliation[AMS]{%
  organization={AMS Institute},
  addressline={Gebouw 027W, Kattenburgerstraat 5},
  city={Amsterdam},
  postcode={1018},
  country={Netherlands}}

\input{frontmatter/abstract}

\end{frontmatter}

%% ----------------------------------------------------------------------
%% Body
%% ----------------------------------------------------------------------

\input{sections/01_introduction}
\input{sections/03_methods_background}
\input{sections/07_results}
\input{sections/08_discussion}

% \input{sections/09_conclusions}

%% ----------------------------------------------------------------------
%% Back matter
%% ----------------------------------------------------------------------

\input{backmatter/credit}
\input{backmatter/declarations}
\input{backmatter/acknowledgements}

%% Bibliography
\printbibliography[heading=bibintoc, title=References]

\end{document}

%% file: frontmatter/abstract.tex
\begin{abstract}
    Police-reported crash statistics remain the standard input for urban road-safety assessment, but their incompleteness and reporting lag limit their usefulness for timely, fine-grained intervention design. Harsh acceleration and braking events are widely used as surrogate safety indicators, but have so far been studied only in comparatively small urban samples. This study analyses harsh events across the urban road network of Milan, combining high-resolution telematics from more than 4.2 million vehicles equipped with On-Board Units, segment-level traffic metrics from TomTom, street-network and infrastructure attributes from OpenStreetMap, and visual streetscape features extracted from Google Street View via semantic segmentation using a OneFormer model. We employ an analytical framework combining non-parametric Mann--Whitney U tests of segment-feature distributions between high- and low-harshness groups with supervised machine-learning regressors. We find that, once exposure is controlled for, wider carriageways, crossings and transit stops, and more open visual fields (higher sky- and road-pixel proportions) are associated with higher harsh-event intensity, while denser built frontage is associated with lower intensity. Finally, the cycling-infrastructure case study identifies a gradient in harsh-event intensity across facility types: markings-only cycle lanes are associated with a 19.5\% higher harshness score, and mixed-traffic configurations with an 11.5\% higher score, relative to physically separated cycle paths, conditional on the included controls. These results support context-specific rather than uniform urban-safety interventions and illustrate how large-scale telematics combined with open geospatial and visual data can inform Vision Zero decision-making at the metropolitan scale.
\end{abstract}

%% file: sections/01_introduction.tex
\section{Introduction and Relevant Literature}
\label{sec:intro}

% \subsection{Urban speed management and crash dynamics}
Urban road safety has gained renewed policy attention, particularly with respect to vulnerable road users such as pedestrians and cyclists. The European Commission's Vision Zero strategy~\parencite{euvisionzero} sets the long-term goal of eliminating traffic fatalities and serious injuries, and is typically implemented through infrastructure upgrades, public awareness activities, and policy reform coupled with enforcement. Several European cities, including Stockholm, Copenhagen, Amsterdam, Z\"urich, and Barcelona, have operationalised this framework through measures such as city-wide 30~km/h speed limits, restrictions on motor-vehicle access, and street redesign~\parencite{yannis2024review}. In Italy, Bologna has recently adopted a 30~km/h limit on most inner streets~\parencite{bologna30}, while similar proposals are being debated in Milan and other cities. The issue is of particular relevance in the Italian context, where traffic fatality rates remain persistently above the European Union average~\parencite{eucommissioncarfatalities} and a substantial share of traffic injuries occurs in urban settings.

Reduced urban speed limits are an effective safety measure: \textcite{yannis2024review} review European experiences with city-wide 30~km/h zones and report an associated 23\% reduction in injury crashes, with no increase in congestion or travel time. Complementary levers such as automated enforcement, vertical calming, credible design~\parencite{sadeghi2016speed, yao2019exploring} and a range of contextual factors (temporal exposure, land use, network connectivity, intersection control~\parencite{cabrera2020uncovering, vorko2006risk, ng2001effects, iranmanesh2024mapping, greibe2003accident}) further shape urban crash risk.

% \subsection{Harsh events as surrogate safety indicators}

Conventional crash-risk evaluations rely almost exclusively on police-reported crashes. Police reports are, however, frequently incomplete, unavailable, or lag by months, which limits their usefulness for timely safety planning. This has motivated a growing interest in \emph{surrogate safety indicators} -- metrics capable of flagging crash risk before crashes occur. Harsh braking and acceleration events (harsh events) are widely used as proxies for aggressive or unsafe driving~\parencite{silva2020systematic} and lend themselves to continuous monitoring through vehicle telematics.

Longitudinal accelerations and decelerations exceeding roughly $1.5$--$2~\mathrm{m\,s^{-2}}$ are routinely treated as indicators of risky manoeuvres in the literature~\parencite{silva2020systematic}. Their spatial structure has been examined by \textcite{ziakopoulos2021spatial} using smartphone trajectory data for an urban suburb of Athens. The authors detected positive spatial autocorrelation of both harsh brakings (HBs) and harsh accelerations (HAs). Geographically Weighted Poisson Regression revealed that segment length and traffic exposure (pass count) are the strongest positive predictors of HBs, while gradient and neighbourhood complexity have negative effects. A follow-up study~\parencite{ziakopoulos2022spatial} confirmed these trends and added that two-lane roads tend to produce more harsh brakings than single-lane segments, while low-volume residential streets produce fewer. Collectively, these analyses demonstrate that harsh events mirror features of the built environment and can therefore support the identification of potentially unsafe locations in advance of recorded crashes. Three data-side limitations nonetheless remain. First, the analysis drew on fewer than 1{,}500 drivers of a single smartphone application, which constrains its spatial coverage and statistical power. Second, the built environment was represented through OSM geometry alone, without visual streetscape information. Third, traffic exposure was controlled through pass counts rather than through an independent measure of traffic volume. The present study addresses these three data-side limitations.

Harsh events in the city of Milan are here analysed using high-resolution telematics records from more than 4.2 million vehicles equipped with On-Board Units are provided by UnipolTech, the telematics branch of one of Italy's largest insurance groups. These data are integrated with segment-level traffic metrics from TomTom, network geometry and infrastructure attributes from OpenStreetMap (snapshot of 31 December 2023), and visual streetscape features extracted from Google Street View (GSV) imagery using a state-of-the-art semantic-segmentation model. The combined dataset covers 24{,}673 urban road segments and 3{,}407{,}876 map-matched harsh events recorded over six representative weeks of 2023.

The analysis combines non-parametric statistical tests, supervised machine-learning regressors with SHAP-based feature attribution, and an exposure-controlled OLS specification for the cycling-infrastructure case study. A dedicated case study on cycling infrastructure examines how different designs are associated with harsh-event intensity on adjacent road segments.

The analysis is organised around three research questions:
\begin{enumerate}
    \item[\textbf{RQ1.}] Do large-scale telematics-based harsh-event measures reproduce the segment-level associations previously reported from smaller smartphone datasets, and does adding visual streetscape descriptors improve their explanatory power? 
    \item[\textbf{RQ2.}] After controlling for traffic exposure, how do road geometry, regulatory attributes, and visual openness relate to harsh-event intensity across the urban network? 
    \item[\textbf{RQ3.}] Can we use harsh events to inform on specific policies decisions, such as cycling infrastructure?
\end{enumerate}

The remainder of the paper is organised as follows. Section~\ref{sec:datamethods} describes the data sources, the processing and integration pipeline, and the methodological background of the statistical and machine-learning models employed. Section~\ref{sec:results} presents the results of the statistical and machine-learning analyses, together with the cycling-infrastructure case study. Section~\ref{sec:discussion} discusses the implications and limitations and outlines directions for future research.

%% file: sections/03_methods_background.tex
\section{Data and Methods}
\label{sec:datamethods}

\subsection{Data sources}
\label{sec:data}

The analysis integrates five complementary datasets: telematics-based harsh events from UnipolTech, segment-level traffic metrics from TomTom, street-network and infrastructure attributes from OpenStreetMap, the Milan Municipal Cycle Network, and Google Street View imagery. Each source and its relevant attributes are described below.

\subsubsection{OpenStreetMap: street network and infrastructure}
\label{sec:data_osm}

Street-network topology and infrastructure attributes were retrieved from \emph{OpenStreetMap} (OSM) for the municipality of Milan, using the \texttt{osmnx} library~\parencite{boeing2025modeling}. The snapshot corresponds to 31 December 2023 and comprises 18{,}037 nodes and 35{,}606 directed edges with a total length of approximately 3{,}640~km and an average segment length of 102~m. For each segment the dataset provides the geometry, number of lanes, street classification (e.g.\ residential, tertiary), posted speed limits (where available), traffic directionality, and the presence of sidewalks and cycleways.

Since the analysis targets urban driving, motorway segments (605 in total) were excluded, together with segments that did not contain any matched UnipolTech observation. The final OSM network comprises 24{,}673 segments.

\begin{figure}[htbp]
    \centering
    \includegraphics[width=0.85\linewidth]{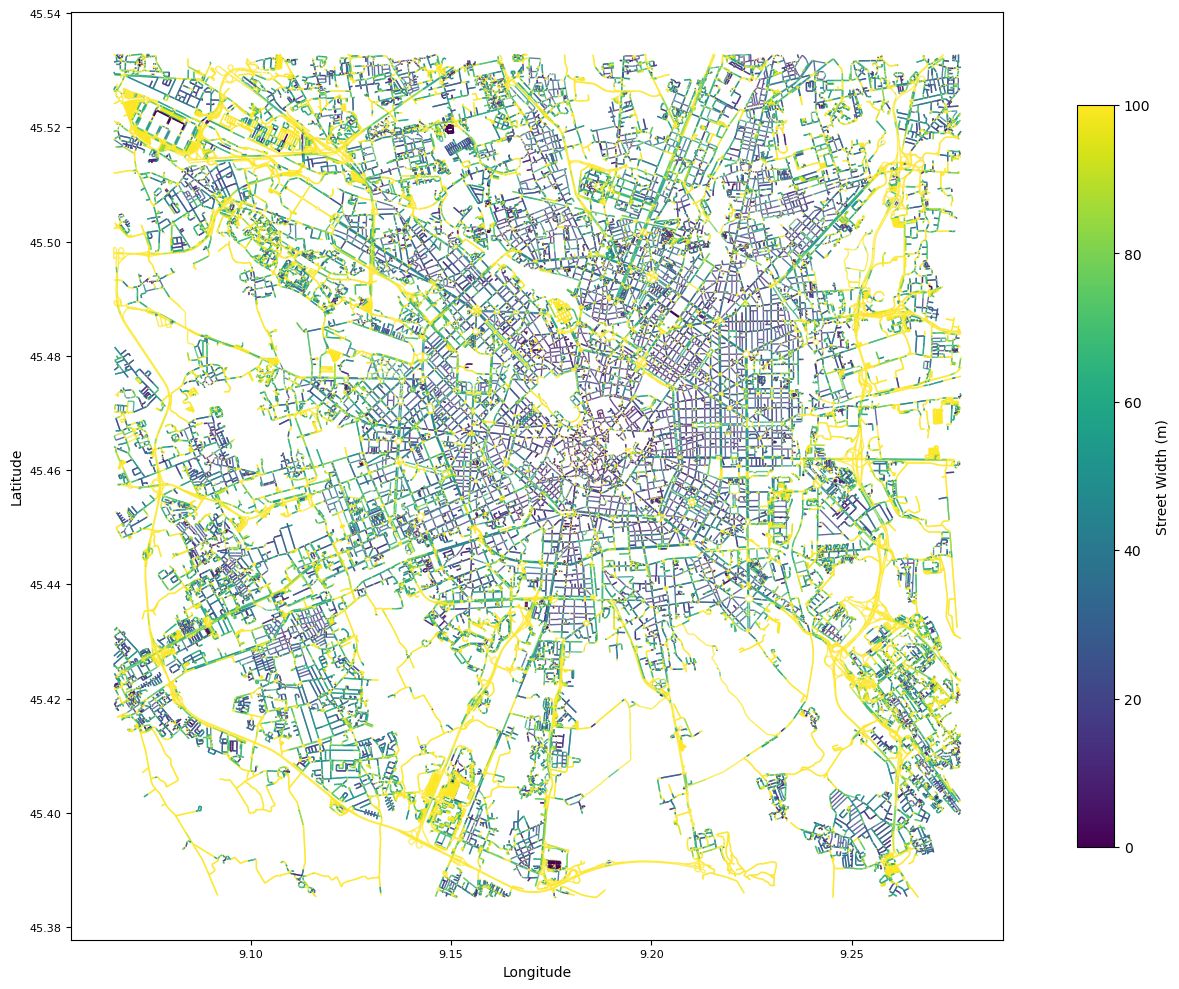}
    \caption{Distribution of estimated street widths across the city of Milan, obtained via the \texttt{momepy.StreetProfile} method.}
    \label{fig:building_width}
\end{figure}

Three feature groups were derived for every OSM segment. First, \emph{geometric} features consist of segment length and sinuosity, defined as
\begin{equation}
\mathrm{sinuosity} = \frac{\mathrm{segment~length}}{\lVert (x_{d}, y_{d}) - (x_{o}, y_{o}) \rVert} ,
\end{equation}
where $(x_{o}, y_{o})$ and $(x_{d}, y_{d})$ denote the coordinates of the origin and destination nodes. Second, \emph{streetscape-context} features count the infrastructure elements within a 20~m buffer around each segment: traffic lights, pedestrian crossings, public-transport stops, cycleways, sidewalks, playgrounds, and green areas. Third, \emph{road form and regulation} features comprise the grouped \texttt{highway} tag, the number of lanes, and an estimated street width computed via the \texttt{momepy.StreetProfile} method~\parencite{Fleischmann2019} following \textcite{araldi2019street}. The distribution of estimated street widths across Milan is shown in Figure~\ref{fig:building_width}.

\subsubsection{UnipolTech: telematics harsh events}
\label{sec:data_unipol}

The primary source of driving-behaviour data is \emph{UnipolTech}, the telematics branch of Unipol -- one of the largest insurance groups in Italy. Data are collected from a nationwide fleet of more than 4.2 million vehicles equipped with On-Board Units (OBUs) that record sudden braking and rapid acceleration events. Each event record contains a timestamp, the event latitude and longitude, instantaneous speed, vehicle heading (degrees from North), the event type (braking or acceleration), a measure of severity, a GNSS signal-quality flag, the device type, and the vehicle category.

Two OBU models are used in the fleet:
\begin{itemize}
    \item \emph{Targa Telematics} devices measure acceleration directly from an on-board accelerometer and assign each event to one of four severity classes, with class boundaries (in $\mathrm{cm\,s^{-2}}$) of 350--450, 450--550, 550--700, and above 700.
    \item \emph{MetaSystem} devices estimate acceleration from GNSS-derived speed changes over a one-second window.
\end{itemize}
The two device families are harmonised onto the Targa class boundaries to yield a unified severity scale from 1 to 4. 

Although the raw dataset covers the whole of Italy, the present study focuses on Milan for three reasons. First, the city offers consistent spatial coverage across all ancillary data sources (OSM, GSV, TomTom). Second, Milan is actively debating Vision Zero--oriented measures such as the extension of 30~km/h zones, which gives the analysis direct policy relevance. Third, restricting the analysis to a single metropolitan area keeps the computation tractable while still yielding several million driving events.

To balance seasonal representativeness against data volume, harsh events were retained for six representative weeks of 2023, corresponding to the first week of January, April, July, October, November, and December.

\label{sec:mapmatching}
Each harsh event was assigned to an OSM segment through a two-step matching procedure. First, every OSM segment was sampled at 3~m intervals; for each sampled point $p_i$ the coordinates $(x_i, y_i)$ and the local tangent heading $h_i$ were computed. For each event, the 20 geographically closest sampled points were retrieved, and the candidate with the smallest absolute heading difference $\lvert h_i - h_{\text{obs}} \rvert$ to the observed vehicle heading was selected. A match was retained only if the spatial distance between the event and the selected point was smaller than 6~m and the heading difference smaller than $45^{\circ}$. These thresholds were calibrated on dense urban layouts to minimise mismatches between parallel carriageways and service roads.

\label{sec:exposure}

Once events were mapped to OSM segments, counts were aggregated at the segment level. To control for exposure, a segment-level exposure metric was computed as
\begin{equation}
\mathrm{Exposure}_{s} = \mathrm{TrafficVolume}_{s} \times \mathrm{Length}_{s} ,
\end{equation}
where $\mathrm{TrafficVolume}_{s}$ is the TomTom probe count for segment $s$ (see Section~\ref{sec:tomtom_matching}) and $\mathrm{Length}_{s}$ is the OSM segment length. Because the raw exposure distribution is right-skewed, a log-transform was applied:
\begin{equation}
\mathrm{LogExposure}_{s} = \log\!\left(1 + \mathrm{Exposure}_{s}\right) .
\end{equation}
The weighted event count on segment $s$ is defined as
\begin{equation}
\label{eq:harshscore}
\mathrm{HarshScore}_{s} = n_{1} + 2\, n_{2} + 3\, n_{3} + 4\, n_{4} ,
\end{equation}
where $n_{k}$ is the number of events of severity level $k$. A normalised, log-transformed harshness rate is further defined as
\begin{equation}
\label{eq:harshrate}
\mathrm{HarshRate}_{s} \;=\; \log\!\left(1 + \dfrac{\mathrm{HarshScore}_{s}}{\mathrm{Exposure}_{s}}\right).
\end{equation}
The harsh score captures the absolute intensity of risky driving on a segment, while the harsh rate captures its per-unit-exposure intensity; both quantities are reported in the following.

\subsubsection{TomTom: segment-level traffic metrics}
\label{sec:data_tomtom}

To control for traffic exposure, segment-level traffic data were obtained from \emph{TomTom}. TomTom aggregates probe data from in-vehicle navigation systems, smartphone applications, and third-party mapping platforms (including Apple Maps). For each of its proprietary segments the dataset provides average and median vehicle speed, posted speed limit, road class and name, speed percentiles (P5, P95), speed standard deviation, and the sample size of contributing probes. The probe sample size is used as a proxy for traffic volume: this is a deliberate simplification, but is supported by TomTom's coverage, which is reported by the data provider to capture signals from a substantial share of the vehicle fleet~\parencite{tomtom_historical_traffic_volumes_2025}.

\label{sec:tomtom_matching}

TomTom uses its own segmentation (mean length 56.1~m), which is finer than OSM (mean length 102.3~m). Matching between the two networks was carried out in a projected metric CRS (EPSG:32632) using an STR-tree spatial index. For each TomTom segment the 20 nearest OSM candidates were retrieved, the Hausdorff distance and the absolute heading difference were computed, and a match was accepted if the minimum Hausdorff distance $d_{\min}$ satisfied $d_{\min} \leq 10$~m and the heading difference $\Delta \theta$ satisfied $\Delta \theta \leq 45^{\circ}$. The directed Hausdorff distance between point sets $A$ and $B$ is
\begin{equation}
h(A, B) = \max_{a \in A} \min_{b \in B} \lVert a - b \rVert ,
\end{equation}
and captures the largest deviation between the two geometries. With these thresholds, all 71{,}867 TomTom segments were matched.

For each TomTom metric $M$ (mean speed, percentiles, speed standard deviation, sample size), the aggregated value for OSM segment $s$ was computed as a length-weighted average:
\begin{equation}
\bar{M}_{s} = \frac{\sum_{t \in \mathcal{T}(s)} L_{t}\, M_{t}}{\sum_{t \in \mathcal{T}(s)} L_{t}} ,
\end{equation}
where $\mathcal{T}(s)$ is the set of TomTom segments matched to $s$ and $L_{t}$ the length of TomTom segment $t$. Longer TomTom segments covering a greater portion of the OSM edge therefore contribute proportionally more to the aggregated estimate.

\subsubsection{City of Milan: municipal cycle network}
\label{sec:data_cycle}

Cycling-infrastructure data were obtained from the \emph{Milan Municipal Cycle Network} via the city's open data portal. Each record provides the geometry and type of the facility (physically separated cycle paths, on-street cycle lanes with markings only, shared-with-vehicles segments, low-traffic zones), the length, the year of construction, and a regulatory classification.
The data was cleaned to remove road crossings and facilities outside the temporal window of the study (keeping only those completed by 2023). Each cycle-path geometry was projected to EPSG:32632 and split into contiguous 3~m fragments, discarding fragments shorter than 10~cm. Each fragment was then matched to its corresponding OSM segment following the procedure of Section~\ref{sec:tomtom_matching}.

\begin{figure}[htbp]
    \centering
    \includegraphics[width=0.70\textwidth]{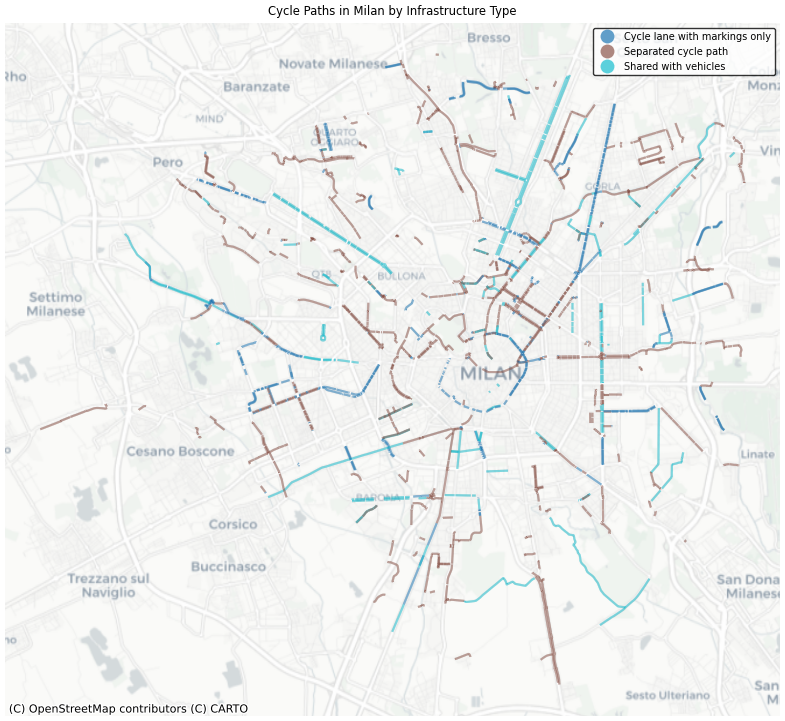}
    \caption{Milan cycling infrastructure by facility type: separated cycle paths, markings-only lanes, and shared-with-vehicles segments.}
    \label{fig:cyclelanes_map}
\end{figure}

Three facility types are compared (Figure~\ref{fig:cyclelanes_map}):
\begin{enumerate}
    \item \emph{Separated cycle path} -- physically segregated from motor-vehicle lanes by curbs, barriers, or grade separation (categories \emph{ciclabile sede propria} and \emph{promiscuo pedoni}).
    \item \emph{Cycle lane with markings only} -- visually delimited lanes on the carriageway, without physical separation (category \emph{ciclabile segnaletica}).
    \item \emph{Shared with vehicles} -- mixed-traffic roads designated for both cyclists and motor vehicles (category \emph{promiscuo veicoli}).
\end{enumerate}

Of the 92{,}058 three-metre cycle-infrastructure fragments, 52\% are markings-only, 25.8\% are separated, and 21.8\% are shared-with-vehicles. After removing fragments without matched exposure or harsh-event data, 84{,}118 fragments entered the analysis.

A log-linear regression was estimated with $\log(\mathrm{HarshScore} + 1)$ as the dependent variable and with $\log(\mathrm{Exposure} + 1)$, posted speed limit, number of lanes, road width, and two facility-type dummies (markings, shared) as predictors. Separated cycle paths are the baseline:
\begin{equation}
\begin{aligned}
\log(\mathrm{HarshScore} + 1)
&= \beta_{0} + \beta_{1} \log(\mathrm{Exposure} + 1) + \beta_{2}\, \mathrm{SpeedLimit} + \beta_{3}\, \mathrm{Lanes} \\
&\quad + \beta_{4}\, \mathrm{RoadWidth} + \beta_{5}\, \mathrm{Markings} + \beta_{6}\, \mathrm{Shared} + \varepsilon .
\end{aligned}
\end{equation}

These data is used to investigate the relationship between cycling-infrastructure design and harsh driving events on the adjacent carriageway.

\subsubsection{Google Street View: visual streetscape context}
\label{sec:data_gsv}

Visual streetscape descriptors were extracted from \emph{Google Street View} (GSV). For each of the 24{,}673 retained OSM segments, two $640\times640$~pixel images were retrieved via the GSV API, one in the direction of travel and one in the opposite direction. GSV was preferred over crowdsourced alternatives such as Mapillary because it provides a standardised, high-resolution imaging pipeline, which is essential for pixel-level comparisons across segments. The 360-degree coverage of each panoramic capture allows arbitrary view orientations to be sampled.

\begin{figure}[htbp]
  \centering
  \begin{subfigure}[t]{0.48\textwidth}
    \centering
    \includegraphics[width=\textwidth]{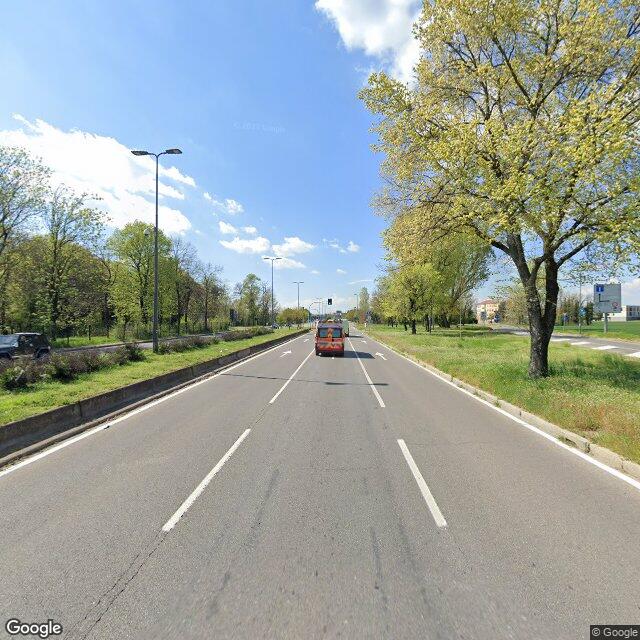}
    \caption{}
    \label{fig:gsv_example_1}
  \end{subfigure}
  \hfill
  \begin{subfigure}[t]{0.48\textwidth}
    \centering
    \includegraphics[width=\textwidth]{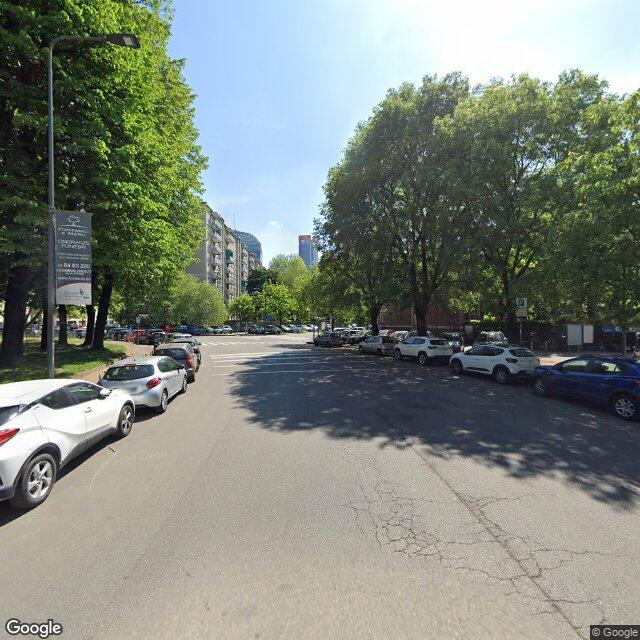}
    \caption{}
    \label{fig:gsv_example_2}
  \end{subfigure}
  \caption{Example Google Street View captures used as input to the semantic-segmentation pipeline.}
\end{figure}

Figures~\ref{fig:gsv_example_1} and~\ref{fig:gsv_example_2} show two example GSV captures used as input to the semantic-segmentation pipeline..

\label{sec:gsv_processing}

\begin{figure}[htbp]
  \centering
  \begin{subfigure}[t]{0.48\textwidth}
    \centering
    \includegraphics[width=\textwidth]{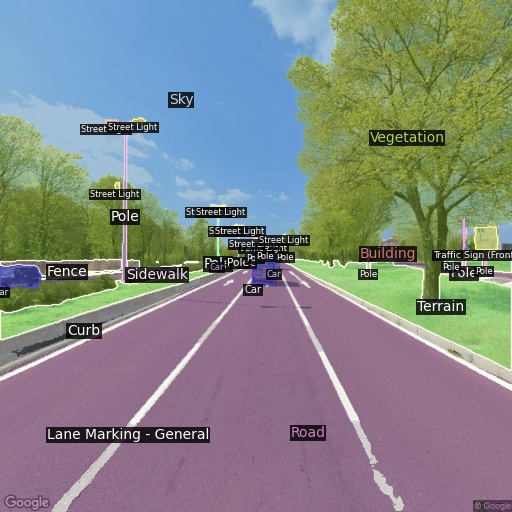}
    \caption{}
    \label{fig:oneformer_example_1}
  \end{subfigure}
  \hfill
  \begin{subfigure}[t]{0.48\textwidth}
    \centering
    \includegraphics[width=\textwidth]{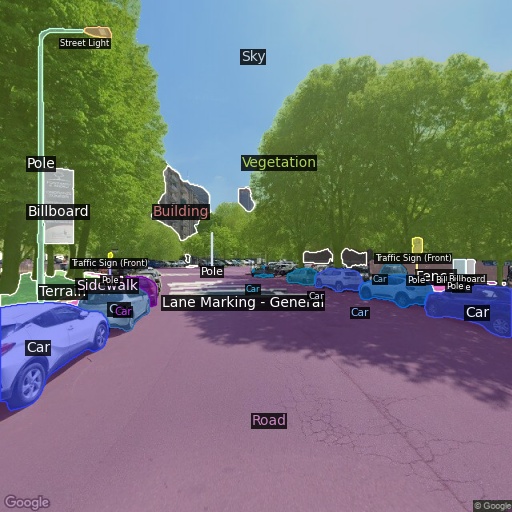}
    \caption{}
    \label{fig:oneformer_example_2}
  \end{subfigure}
  \caption{Two examples of GSV captures segmented by OneFormer into eleven super-classes.}
\end{figure}

Segmentation was performed with \emph{OneFormer}~\parencite{jain2022oneformer}, equipped with a DiNAT-L backbone pretrained on Mapillary Vistas v2.0 (47.8 PQ, 64.0 mIoU on the benchmark). The raw Mapillary categories were aggregated into eleven super-classes: road, sidewalk, building, vegetation, sky, street furniture, vehicles, pedestrians, signs, rails, and water. Two examples of segmented images are shown in Figures~\ref{fig:oneformer_example_1} and~\ref{fig:oneformer_example_2}.

For each segmented image a feature vector was built comprising (i) the proportion of pixels belonging to each super-class and (ii) the count of detected objects per class. The two images associated with a segment were subsequently averaged to produce a single segment-level visual descriptor vector.

\subsection{Methods}
\label{sec:methods}

Three complementary analytical approaches are applied to the segment-level dataset.

\emph{Non-parametric comparison.} Segments are split at the median harsh score and distributions of OSM- and GSV-derived features are compared between the high- and low-harshness groups using two-sided Mann--Whitney U tests. For each feature the difference of means and the two-sided $p$-value are reported. The test is used as a directional screen rather than a formal effect-size estimator; at the per-group sample sizes available here even small distributional differences register as statistically significant, and multiple-comparison correction is not applied across the several dozen features compared (Section~\ref{sec:discussion}).

\emph{Machine-learning regression.} Segment-level harsh scores are predicted from geometric, regulatory, traffic, and visual descriptors using Gradient Boosting and Random Forest regressors, trained on an 80\% / 20\% train--validation split with hyperparameters selected by five-fold cross-validated grid search. Predictive performance is reported through $R^{2}$, RMSE, MAE, and MAPE on the held-out set. Feature importance is inspected through classical Gini-based rankings and SHapley Additive exPlanations (SHAP) values, the latter providing a consistent attribution of each prediction to its input features while accounting for interactions. The interaction between exposure as a predictor and exposure as a denominator in the harsh rate is discussed in Section~\ref{sec:discussion}.

\emph{Cycling-infrastructure OLS.} The case study of Section~\ref{sec:cycle_case} estimates a log-linear regression of $\log(\mathrm{HarshScore}+1)$ on $\log(\mathrm{Exposure}+1)$, posted speed limit, number of lanes, road width, and two facility-type dummies, with HC3 robust standard errors. The facility-type coefficients admit a direct percent-change interpretation relative to the separated-cycle-path baseline.

%% file: sections/07_results.tex
\section{Results}
\label{sec:results}

\subsection{Spatial and temporal distribution of harsh events}
\label{sec:results_spatiotemporal}

\begin{figure}[htbp]
    \centering
    \includegraphics[width=0.95\textwidth]{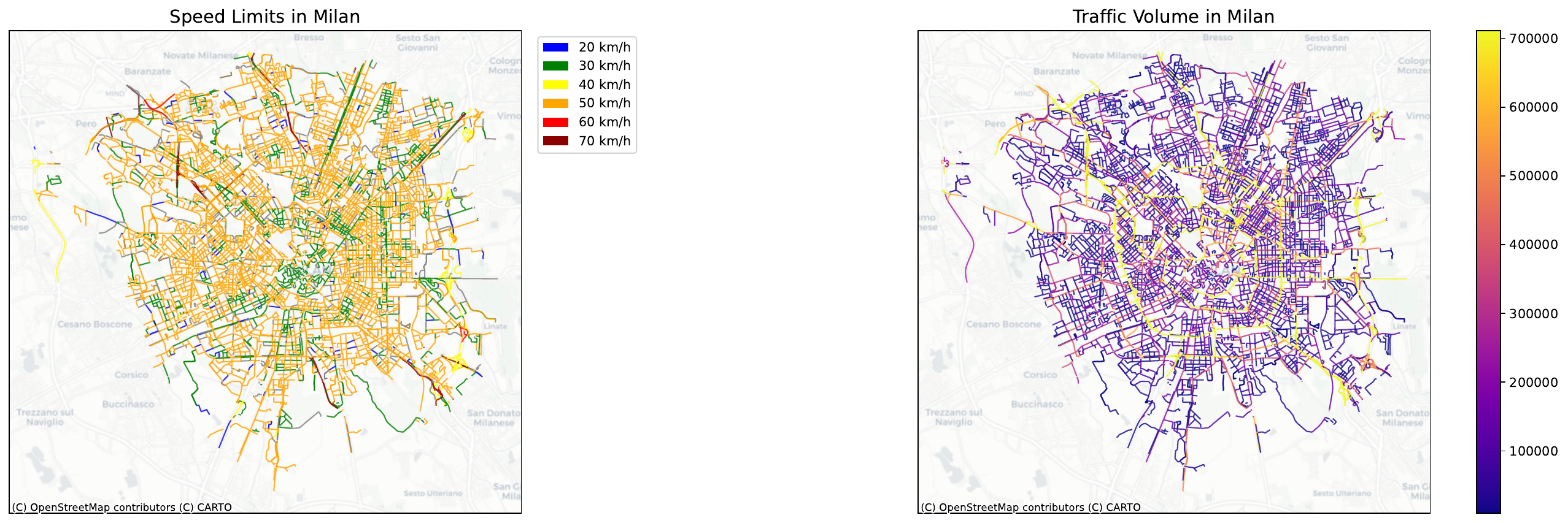}
    \caption{Posted speed limits (left) and TomTom-derived traffic volume proxy (right) across Milan.}
    \label{fig:speed_traffic_milan}
\end{figure}

Posted speed limits in Milan cluster around three values -- 30, 50, and 70~km/h -- which together account for the vast majority of segments. Localised 20~km/h zones exist for pedestrian safety, while 40~km/h limits are reserved for highway ramps and limits above 90~km/h for motorways outside the study area. Figure~\ref{fig:speed_traffic_milan} maps speed limits and TomTom probe counts across the city.

\begin{figure}[htbp]
    \centering
    \includegraphics[width=0.65\textwidth]{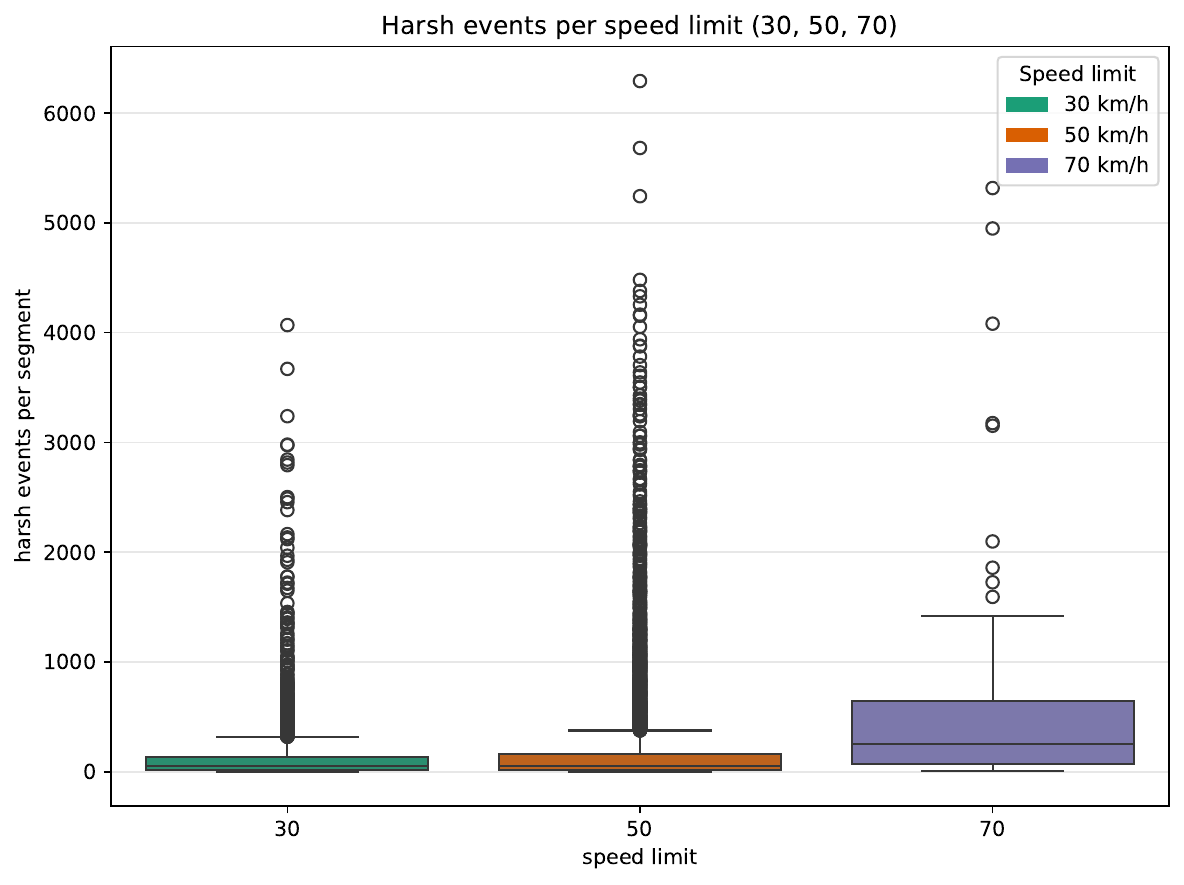}
    \caption{Distribution of harsh scores per segment across the three dominant speed-limit classes.}
    \label{fig:harsh_per_speed}
\end{figure}

\begin{figure}[htbp]
    \centering
    \includegraphics[width=0.65\textwidth]{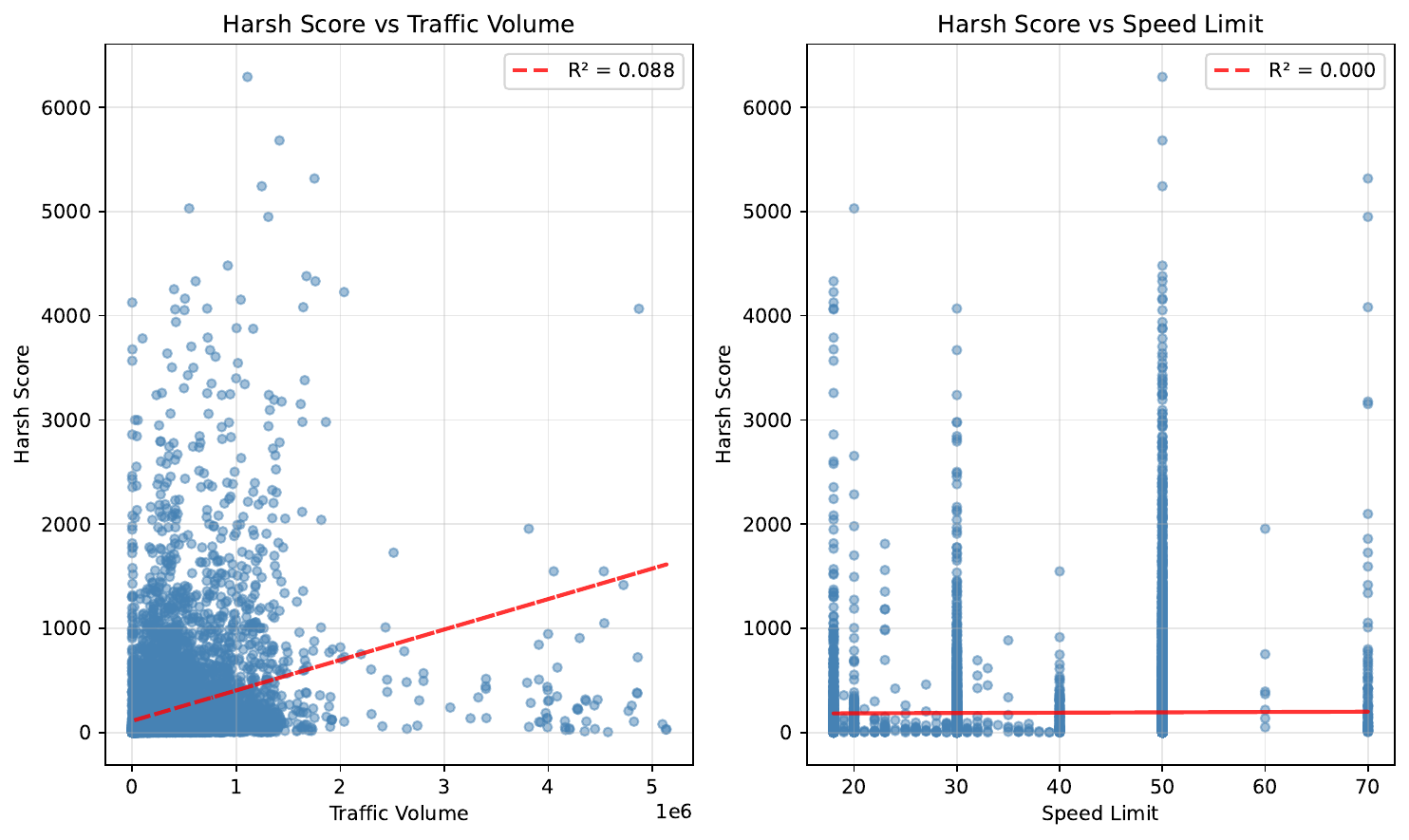}
    \caption{Univariate relationships between the harsh score and traffic volume (left) and speed limit (right).}
    \label{fig:harsh_vs_speed_traffic}
\end{figure}

A direct comparison of harsh events across the three primary speed-limit categories (Figure~\ref{fig:harsh_per_speed}) shows that the absolute number of events increases with the speed limit, with the highest counts observed on 70~km/h roads. This is consistent with higher traffic volumes on those arteries. However, the univariate relationship between the harsh score and either the speed limit or traffic volume alone is weak, with near-zero coefficients of determination ($R^{2} < 0.001$ and $R^{2} = 0.088$; Figure~\ref{fig:harsh_vs_speed_traffic}). Univariate proxies are therefore insufficient and multivariate analyses are required.

\begin{figure}[htbp]
    \centering
    \includegraphics[width=0.95\textwidth]{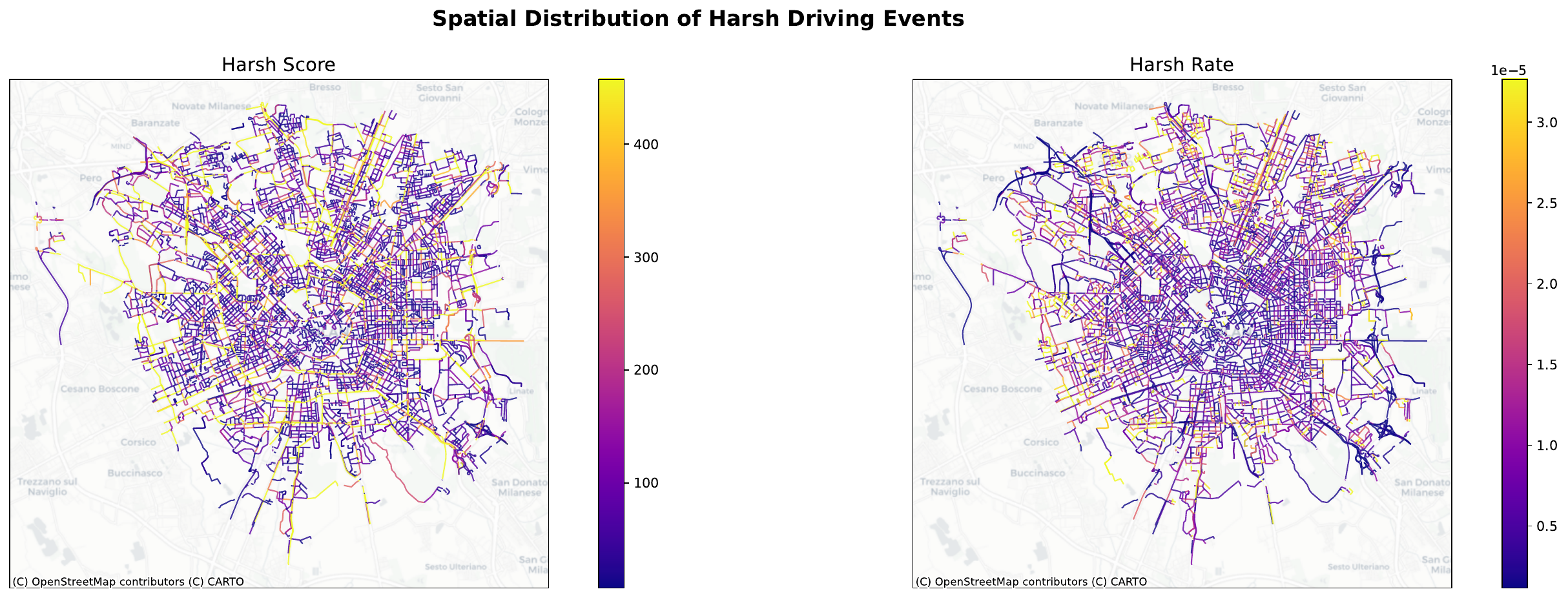}
    \caption{Spatial distribution of the absolute harsh score (left) and the exposure-normalised harsh rate (right) across Milan.}
    \label{fig:spatial_distribution}
\end{figure}

Accounting for exposure reveals complementary spatial patterns (Figure~\ref{fig:spatial_distribution}). The harsh-score map highlights major arterials as the most affected corridors, reflecting high traffic volumes. The harsh-rate map, by contrast, displays more localised hotspots and suggests that certain street geometries and design features drive per-vehicle risk independently of volume.

\begin{figure}[htbp]
    \centering
    \includegraphics[width=0.95\textwidth]{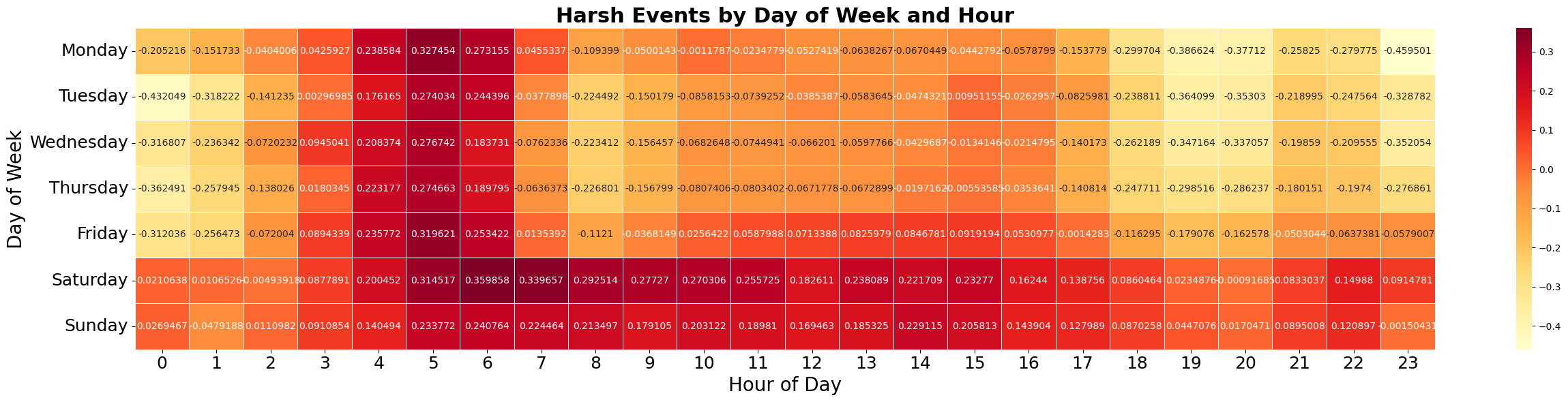}
    \caption{Hourly distribution of the absolute harsh score, by day of the week.}
    \label{fig:harsh_score_hour}
\end{figure}

\begin{figure}[htbp]
    \centering
    \includegraphics[width=0.95\textwidth]{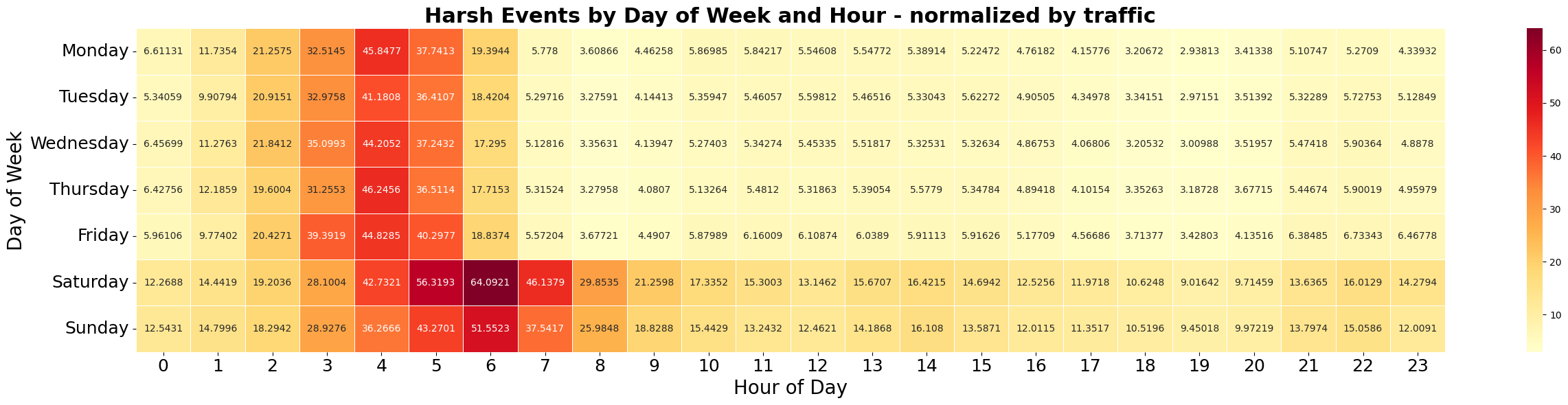}
    \caption{Hourly distribution of the exposure-normalised harsh rate, by day of the week.}
    \label{fig:harsh_rate_hour}
\end{figure}

Temporal patterns echo this distinction. The harsh score peaks during morning and evening commute hours, with particularly pronounced spikes on Monday mornings and Friday afternoons (Figure~\ref{fig:harsh_score_hour}). Normalising for exposure reverses the picture: the harsh rate peaks during night-time hours, especially on weekends (Figure~\ref{fig:harsh_rate_hour}).

\begin{figure}[htbp]
    \centering
    \includegraphics[width=0.60\textwidth]{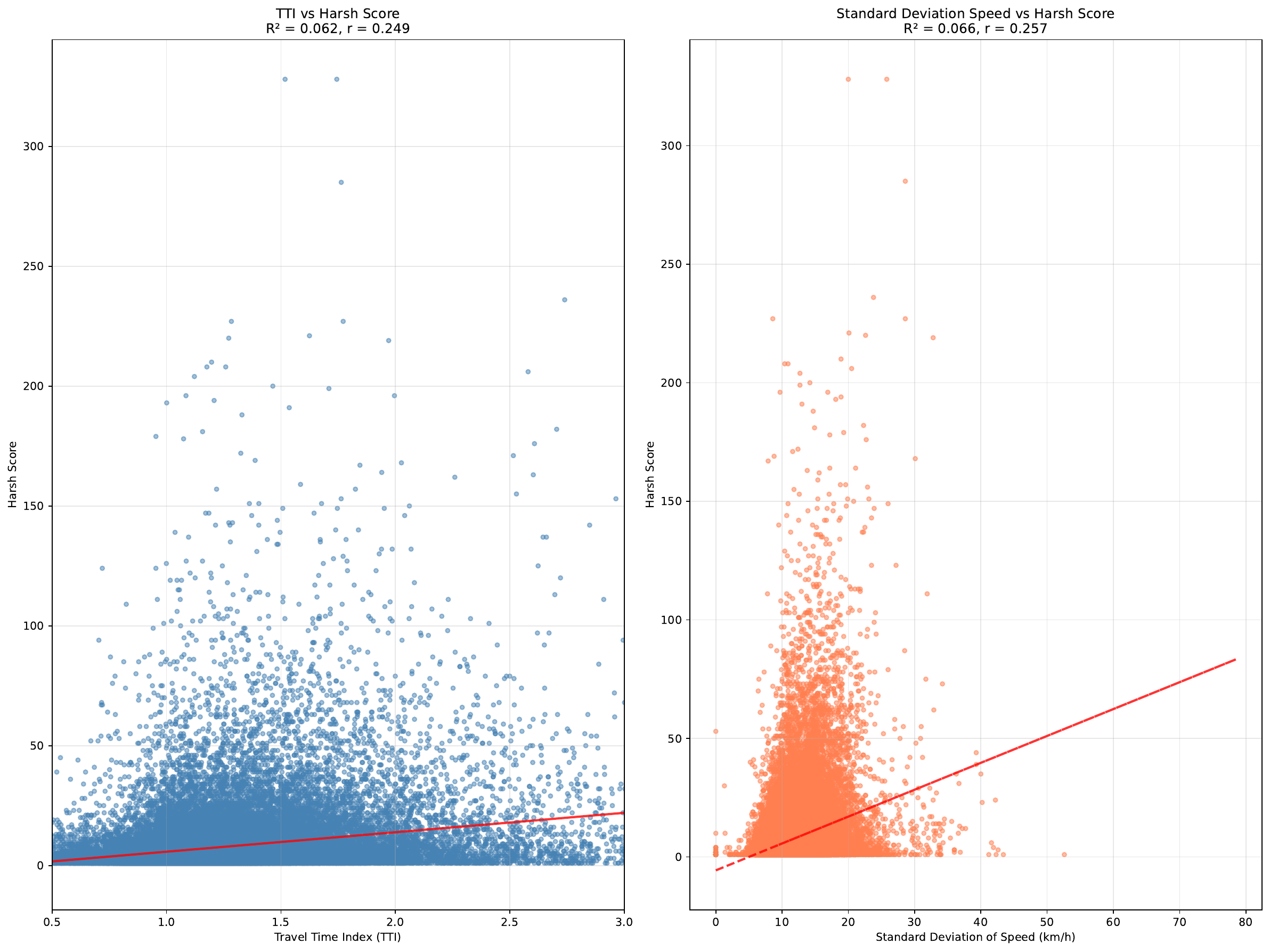}
    \caption{Univariate relationships between the harsh score and the Travel Time Index and speed standard deviation.}
    \label{fig:harsh_vs_congestion}
\end{figure}

Congestion metrics alone are similarly insufficient. Fitting a naive univariate model on the speed standard deviation or on the Travel Time Index (TTI), defined as
\begin{equation}
\mathrm{TTI} = \frac{t_{\text{travel}}}{t_{\text{free-flow}}} ,
\end{equation}
yields $R^{2} = 0.066$ and $R^{2} = 0.062$ respectively (Figure~\ref{fig:harsh_vs_congestion}), again motivating a multivariate treatment.

\subsection{Mann--Whitney U comparisons of segment features}
\label{sec:results_mwu}

\begin{figure}[htbp]
    \centering
    \includegraphics[width=0.60\textwidth]{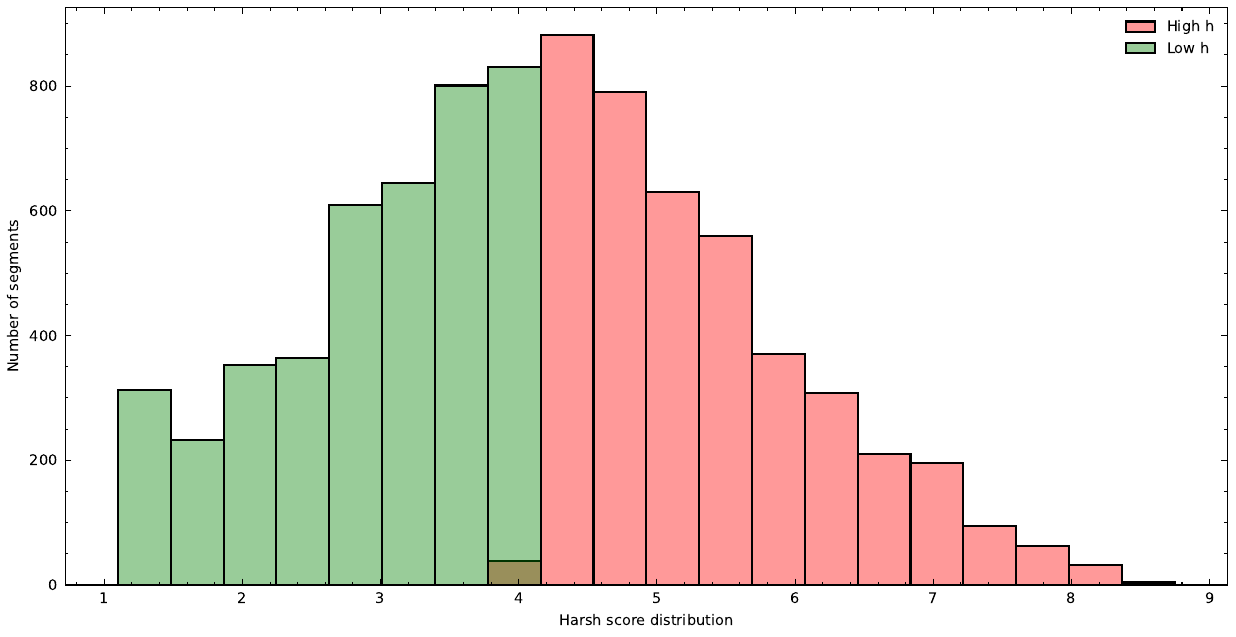}
    \caption{Distribution of the log harsh score across segments.}
    \label{fig:harsh_distribution}
\end{figure}

For the 9{,}267 segments for which GSV imagery was available, segments were split at the median harsh score (logarithm shown in Figure~\ref{fig:harsh_distribution}) and the distributions of OSM- and GSV-derived features were compared between the high- and low-harshness groups using two-sided Mann--Whitney U tests. For each feature $f$, the difference of means $d_{f} = \mu_{f}^{\text{high}} - \mu_{f}^{\text{low}}$ is reported together with the two-sided $p$-value. The analysis is restricted to segments with a 30 or 50~km/h speed limit. 

% At the available per-group sample size ($n \approx 4{,}633$) the test is powerful enough to flag very small distributional differences as statistically significant, and no multiple-comparison correction is applied across the several dozen features compared; the reported $p$-values and differences of means should therefore be interpreted jointly, and as directional indicators rather than effect magnitudes. Formal effect-size measures (Cliff's delta, rank-biserial correlation) are not reported in the current analysis (Section~\ref{sec:discussion}).

\begin{figure}[htbp]
    \centering
    \includegraphics[width=0.95\textwidth]{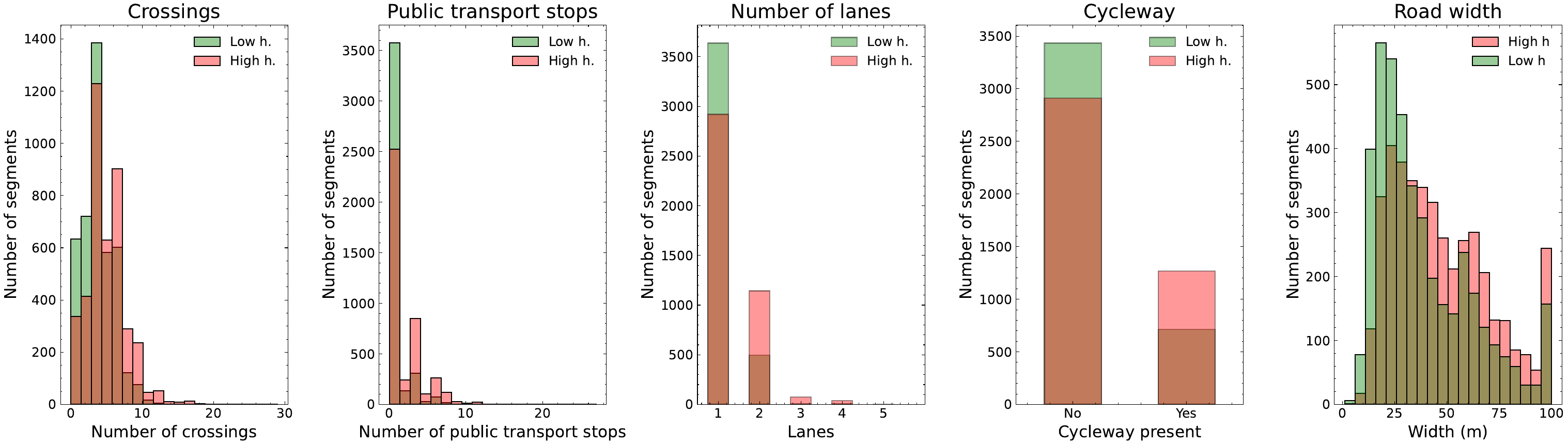}
    \caption{Mann--Whitney U comparisons between high- and low-harshness groups for OSM-derived features.}
    \label{fig:harsh_osm}
\end{figure}

\begin{figure}[htbp]
    \centering
    \includegraphics[width=0.95\textwidth]{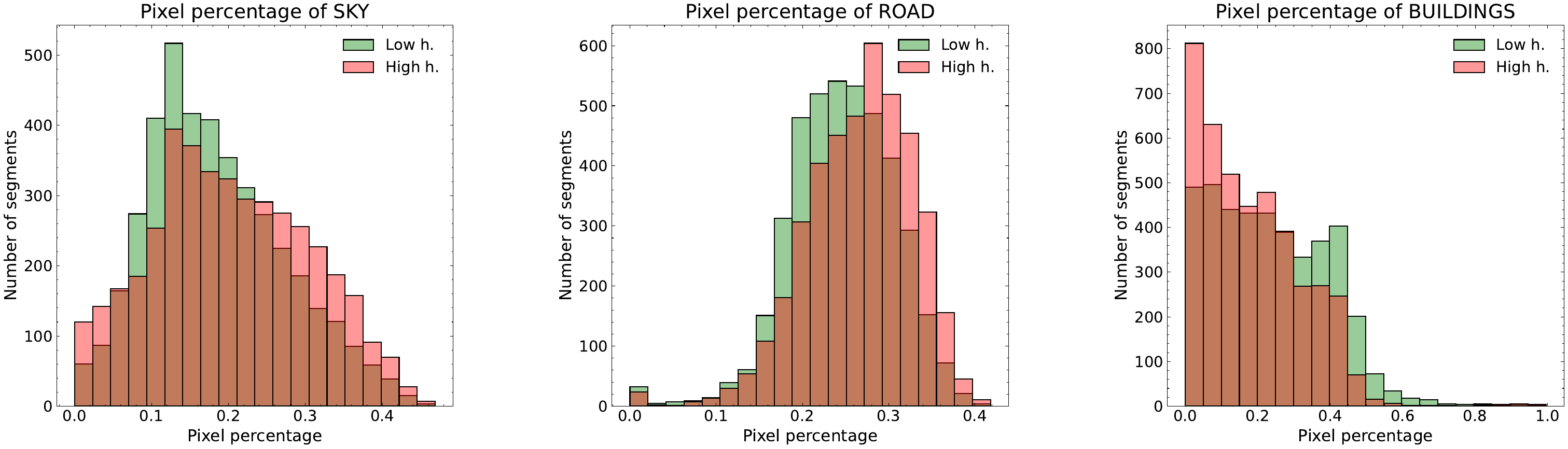}
    \caption{Mann--Whitney U comparisons between high- and low-harshness groups for GSV-derived features.}
    \label{fig:harsh_gsv}
\end{figure}

Road-design and infrastructure attributes differ systematically between the two groups (Figure~\ref{fig:harsh_osm}). Segments with pedestrian crossings ($d = +1.20$), public-transport stops ($d = +1.20$), additional lanes ($d = +0.21$), binary cycleway presence ($d = +0.13$), and greater road width ($d = +9.13$~m) all tend to appear more often in the high-harshness group. GSV-derived visual features show the same direction (Figure~\ref{fig:harsh_gsv}): higher proportions of sky ($d = +0.02$) and road surface ($d = +0.02$) pixels co-occur with higher harsh scores, whereas a higher proportion of building pixels ($d = -0.05$) co-occurs with lower harsh scores.

\subsection{Machine-learning models}
\label{sec:results_ml}

Two feature configurations were compared: a design-only baseline (Model~A) using geometric, regulatory, traffic, and visual descriptors, and a temporal-augmented configuration (Model~B) adding the \texttt{time\_of\_day} and \texttt{is\_week} predictors. Two tree-ensemble algorithms -- Gradient Boosting and Random Forest -- were trained for each configuration, tuned by grid search with five-fold cross-validation. Gradient Boosting was the best-performing algorithm in the design-only configuration ($R^{2} = 0.565$ on the held-out validation set), while Random Forest was the best performer in the temporal-augmented configuration ($R^{2} = 0.714$). Table~\ref{tab:model_performance} therefore reports the best algorithm per configuration.

\begin{table}[htbp]
\centering
\caption{Predictive performance on the held-out validation set. Each row reports the best-performing tree-ensemble algorithm for the corresponding feature configuration.}
\label{tab:model_performance}
\begin{tabular}{lcccc}
\toprule
\textbf{Model} & \textbf{RMSE} & \textbf{MAE} & \textbf{MAPE (\%)} & \textbf{$R^{2}$} \\
\midrule
Design-only (Gradient Boosting) & 0.9867 & 0.7691 & 24.60 & 0.565 \\
With temporal features (Random Forest) & 0.4763 & 0.3689 & 29.80 & 0.714 \\
\bottomrule
\end{tabular}
\end{table}

Adding temporal predictors reduces RMSE and MAE and raises $R^{2}$ from 0.565 to 0.714.

% while MAPE rises from 24.6\% to 29.8\%; the MAPE increase is driven by larger relative errors on low-activity segments (long tail of the harsh-score distribution).

\begin{figure}[htbp]
    \centering
    \begin{subfigure}[t]{0.95\textwidth}
        \centering
        \includegraphics[width=\textwidth]{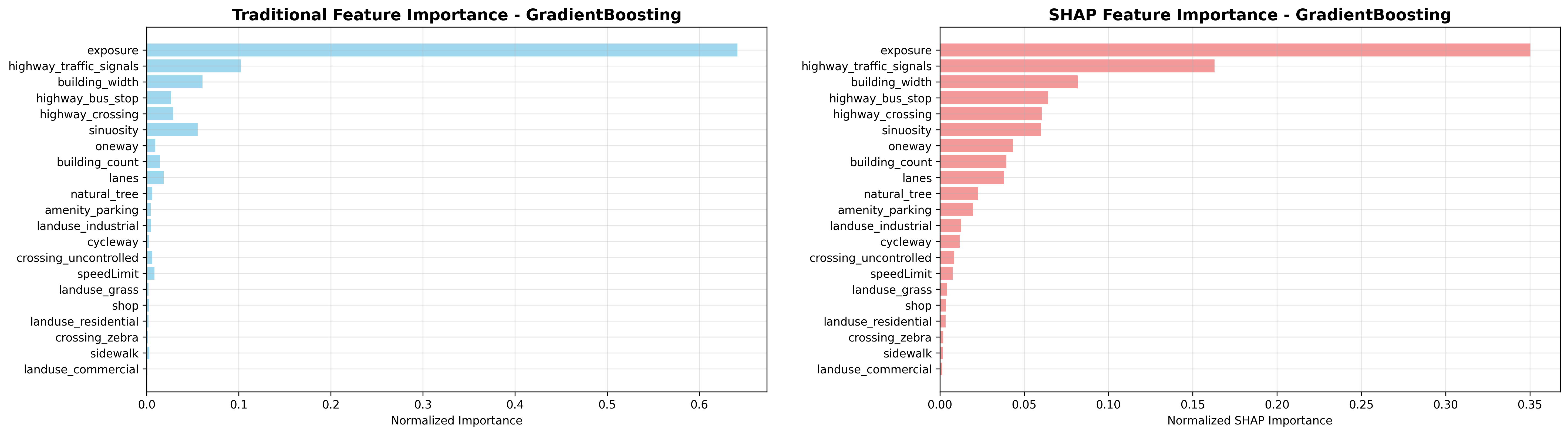}
        \caption{Gini vs.\ SHAP feature-importance rankings, design-only Gradient Boosting model.}
        \label{fig:feature_importance_notime}
    \end{subfigure}

    \vspace{0.5em}

    \begin{subfigure}[t]{0.48\textwidth}
        \centering
        \includegraphics[width=\textwidth]{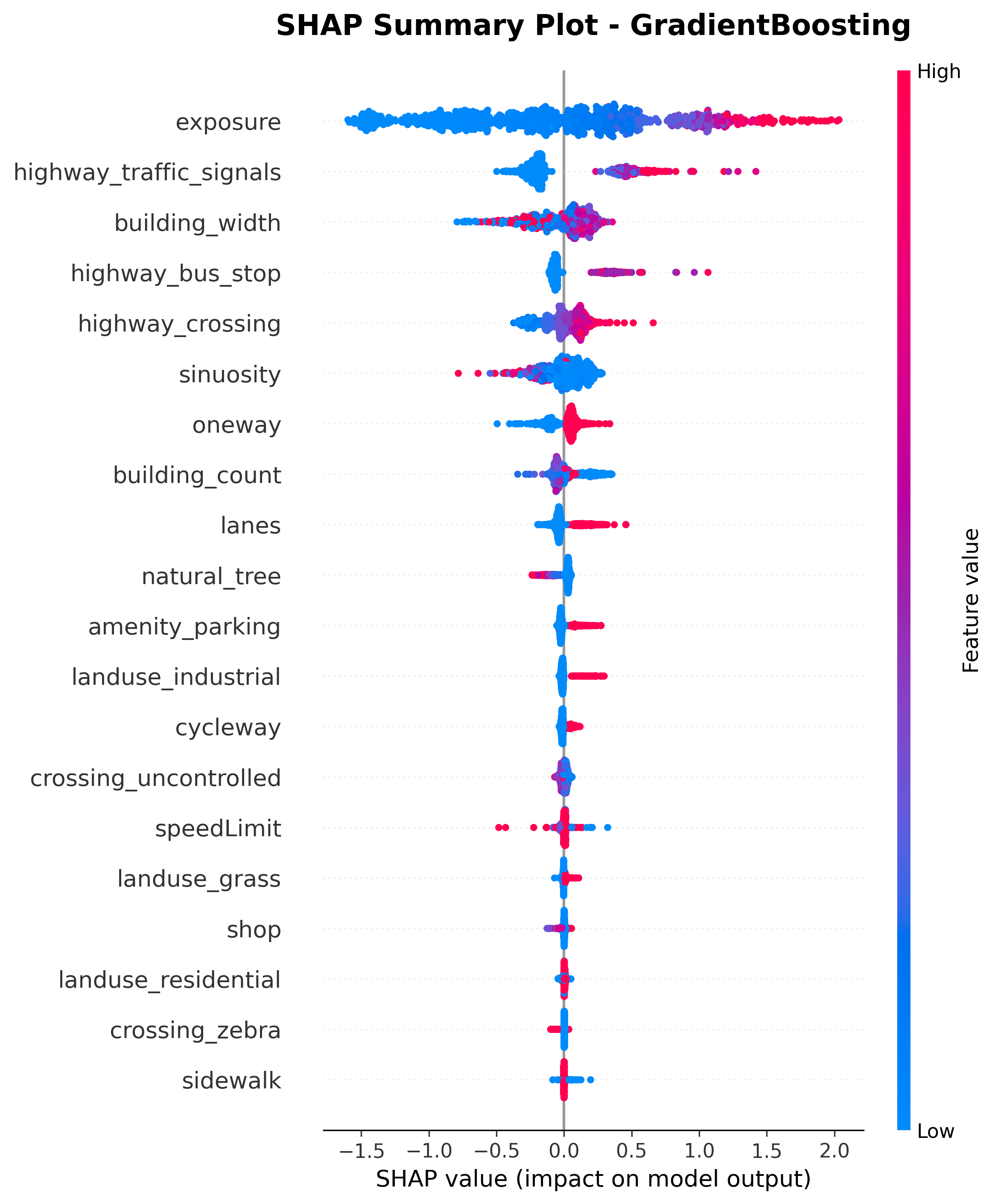}
        \caption{SHAP summary, design-only (Gradient Boosting).}
        \label{fig:shap_summary_notime}
    \end{subfigure}
    \hfill
    \begin{subfigure}[t]{0.48\textwidth}
        \centering
        \includegraphics[width=\textwidth]{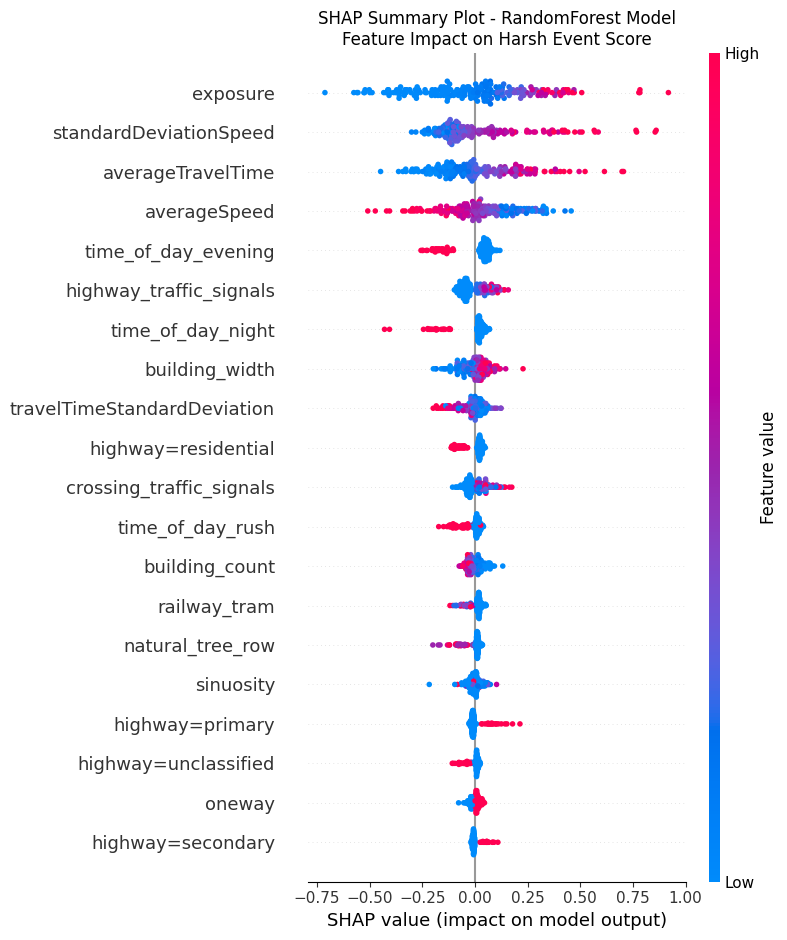}
        \caption{SHAP summary, temporal-augmented (Random Forest).}
        \label{fig:shap_time}
    \end{subfigure}
    \caption{Feature attribution for the two machine-learning regressors. (a) Gini vs.\ SHAP rankings for the design-only Gradient Boosting model. (b) SHAP summary for the design-only model. (c) SHAP summary for the temporal-augmented Random Forest model.}
    \label{fig:shap_combined}
\end{figure}

SHAP-based feature attribution is shown for both models in Figure~\ref{fig:shap_combined}. In the design-only model, segment-level exposure ranks as the single most influential predictor, followed by the number of traffic signals, estimated street width, and the presence of bus stops and pedestrian crossings. In the temporal-augmented model, temporal indicators join the top predictors without displacing exposure.

\subsection{Case study: cycling infrastructure}
\label{sec:cycle_case}

As a targeted application of the framework, we examined whether different designs of cycling infrastructure are associated with differences in harsh driving behaviour on the adjacent carriageway.

The facility-type dummies admit a direct percent-change interpretation: $\log(Y_{\text{treatment}} / Y_{\text{baseline}}) = \beta$, so the proportional change in the expected harsh score relative to the baseline is $(e^{\beta} - 1) \times 100\%$.

Table~\ref{tab:ols_cycle_only} reports the fitted coefficients with HC3 robust 95\% confidence intervals. The model achieves $R^{2} = 0.394$.

\begin{table}[htbp]
\centering
\caption{OLS on $\log(\mathrm{HarshScore} + 1)$ by cycle-infrastructure type, with robust (HC3) 95\% confidence intervals.}
\label{tab:ols_cycle_only}
\small
\begin{tabular}{lccc}
\toprule
Variable & Coef.\ ($\beta$) & 95\% CI low & 95\% CI high \\
\midrule
$\log(\mathrm{Exposure} + 1)$ & $0.584^{***}$ & 0.576 & 0.592 \\
Markings only (vs.\ separated) & $0.178^{***}$ & 0.156 & 0.201 \\
Shared with vehicles (vs.\ separated) & $0.109^{***}$ & 0.089 & 0.130 \\
\midrule
\multicolumn{4}{l}{\emph{Percent change in $(y+1)$}} \\
Markings only & 19.5\% & 16.9\% & 22.3\% \\
Shared with vehicles & 11.5\% & 9.3\% & 13.9\% \\
\midrule
Observations & \multicolumn{3}{c}{84{,}118} \\
Adjusted $R^{2}$ & \multicolumn{3}{c}{0.394} \\
\bottomrule
\multicolumn{4}{p{0.85\linewidth}}{\footnotesize $^{***}p<0.001$. Facility-type effects are reported as $100\,(e^{\beta}-1)$. The coefficient on $\log(\mathrm{Exposure} + 1)$ is an elasticity of $(y+1)$ with respect to $(\mathrm{Exposure}+1)$.}
\end{tabular}
\end{table}

\begin{figure}[htbp]
    \centering
    \includegraphics[width=0.80\textwidth]{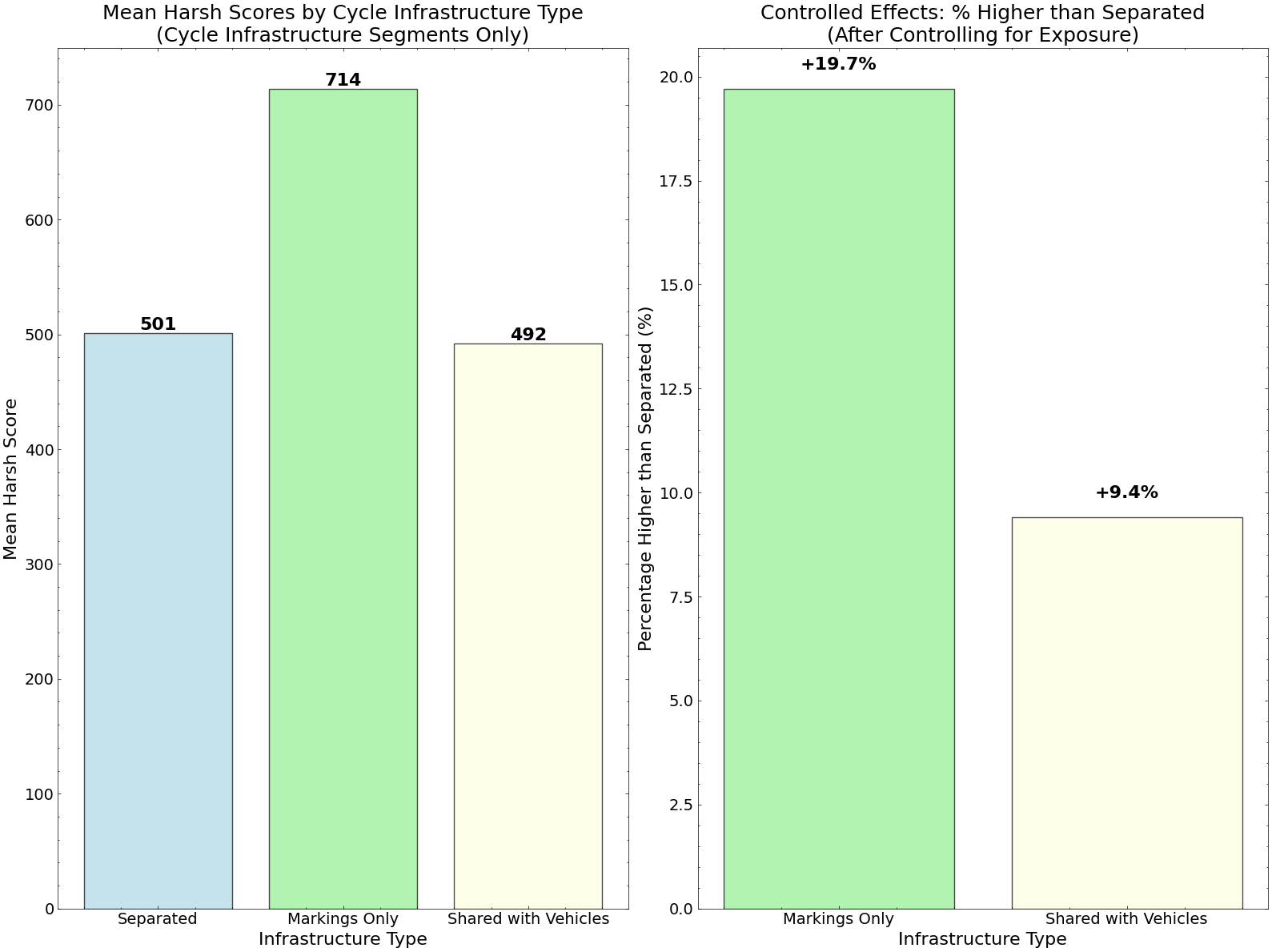}
    \caption{Fitted percent-change in the harsh score by cycling-infrastructure type, relative to separated cycle paths.}
    \label{fig:cyclelanes_results}
\end{figure}

Relative to separated cycle paths, markings-only lanes are associated with a 19.5\% higher harsh score ($p < 0.001$) and shared-with-vehicles configurations with an 11.5\% higher harsh score ($p < 0.001$). The direction of the gradient is consistent across exposure quintiles and is summarised graphically in Figure~\ref{fig:cyclelanes_results}.

%% file: sections/08_discussion.tex
\section{Discussion and Conclusions}
\label{sec:discussion}
This study presents a large-scale spatial analysis of harsh driving events in the urban road network of Milan, using high-resolution telematics data from more than 4.2~million vehicles together with open geospatial sources (OpenStreetMap, TomTom) and visual streetscape features extracted from Google Street View via semantic segmentation. A multi-scale analytical framework (Mann--Whitney U tests and supervised tree-ensemble regressors with SHAP-based feature attribution) was applied to 24{,}673 OSM segments in the metropolitan area. The framework extends the spatial harsh-event analysis of \textcite{ziakopoulos2021spatial} to an order of magnitude larger sample, adds visual streetscape descriptors, and introduces an independent exposure control.

The results converge on two main findings. First, exposure sets the absolute count baseline for harsh-event frequency; once exposure is controlled, segment form and visual openness also exert significant effects. Second, temporal normalisation matters: absolute harsh scores peak during commute hours, while per-vehicle risk peaks on weekend nights. Finally the cycling-infrastructure case study illustrates the framework's practical applicability.

Beyond the quantitative findings, the study contributes a reproducible methodology for rapid construction and analysis of urban road-network spatial datasets that integrate large-scale telematics with open geospatial and visual sources.

\subsection{Interpretation of the spatial, temporal, and design patterns}

The univariate analyses confirm that no single attribute, speed limit, traffic volume, congestion index, explains the spatial distribution of harsh events on its own. The near-zero coefficients of determination reported in Section~\ref{sec:results_spatiotemporal} motivate the multivariate Mann--Whitney and machine-learning analyses, which together indicate that exposure, street form, and visual openness jointly shape the observed spatial pattern. Exposure sets the absolute count baseline for harsh-event frequency; once exposure is controlled, the design and temporal features drive residual variation. This interpretation is consistent with the SHAP rankings, in which exposure is the single most influential predictor but is accompanied at the top of the list by infrastructure attributes (traffic signals, street width, bus stops, pedestrian crossings) and, in the temporal model, by the time-of-day and weekend indicators.

The temporal pattern is itself informative. Absolute harsh scores peak during morning and evening commute hours on weekdays, consistent with commuting stress and traffic density during these periods. Exposure-normalised rates peak at night, especially on weekends, a pattern consistent with reduced visibility, driver fatigue, and alcohol-related impairment reported in the wider crash literature~\parencite{vorko2006risk}. The divergence between the absolute and normalised temporal profiles underscores the importance of reporting both quantities: policies driven by absolute counts will target commute hours, whereas policies driven by per-vehicle risk will target weekend nights.

The Mann--Whitney comparisons and SHAP attributions point in the same direction on road design. Wider carriageways, longer sightlines, more open visual fields (higher sky and road-pixel proportions), and public-transport and pedestrian infrastructure co-occur with higher harsh scores, while denser built frontage and narrower cross-sections co-occur with lower harsh scores. These are associations rather than causal effects, but they are consistent with the broader streetscape-perception literature~\parencite{fan2023urban,li2022street} and suggest that the openness of the driver's visual field is a plausible contributor to aggressive acceleration and braking.

% \subsection{Circularity between exposure and the harshness outcome}

% Two related but distinct exposure specifications enter this study. The exposure-normalised \emph{harsh rate} (Eq.~\ref{eq:harshrate}) uses exposure as a denominator, whereas the machine-learning regressors described in Section~\ref{sec:results_ml} use \texttt{LogExposure} as a predictor of the absolute log harsh score. These parameterisations address different questions: the rate captures per-unit-exposure risk, while the predictor form quantifies how much exposure explains of the absolute count. Because the absolute count is mechanically increasing in exposure (more vehicles produce more events), exposure is expected to dominate SHAP rankings by construction; the ML results should therefore be read as describing how residual variation, once exposure is controlled, relates to design and temporal features, rather than as a free comparison between exposure and design effects.

\subsection{Cycling-infrastructure case study}

The cycling-infrastructure comparison recovers an observed gradient in harsh-event intensity across facility types, conditional on the included controls (exposure, posted speed limit, number of lanes, road width). Markings-only lanes are associated with a 19.5\% higher harsh score and shared-with-vehicles configurations with an 11.5\% higher harsh score, both relative to separated cycle paths. These associations are consistent with the direction suggested by the wider cycling-safety literature but do not on their own establish a causal ranking of facility types. Cycle-path type is highly correlated with road hierarchy, neighbourhood context, and demand patterns: separated paths in Milan tend to be on wider higher-class roads or peripheral greenway corridors, while markings-only and shared facilities appear more frequently on smaller streets in dense districts. The 6-predictor OLS specification reaches $R^{2} = 0.394$, which leaves substantial heterogeneity unmodelled. 
% Neighbourhood fixed effects and road-class stratification were not applied here; the reported coefficients should therefore be read as associations within this specific sample rather than as a causal ranking of design types, and any infrastructure-policy implication would require further causal investigation.

\subsection{Implications for urban-safety interventions}

Taken together, the patterns reported above support context-specific rather than uniform safety interventions. Streets combining wide cross-sections, open sightlines, and unobstructed visual fields emerge as candidate locations for lane narrowing, visual or physical traffic-calming elements, and added roadside built frontage. The temporal patterns suggest that time-of-day enforcement or signal retiming may be appropriate on streets that concentrate per-vehicle risk in the late-night or weekend hours. The cycling-infrastructure gradient is consistent with prioritising physical separation where feasible and, where physical separation is impractical, preferring explicit shared arrangements over paint-only delimitation, subject to further causal validation.

\subsection{Limitations}
\label{sec:limitations}
% Several limitations qualify the interpretation of these results.

The analysis is observational and cannot establish causality. Unmeasured confounders -- enforcement intensity, signal coordination, socio-economic context, cyclist volumes, land-use mix -- may drive part of the observed patterns. The OSM snapshot is fixed to 31 December 2023, but Google Street View capture dates vary across locations; transient factors such as weather, roadworks, or special events are not modelled. The probe-count proxy for traffic volume, although derived from a large and representative sample, remains a simplified measure of exposure.

The telematics fleet combines two device families with different noise characteristics: Targa devices record accelerometer-derived events, whereas MetaSystem devices derive events from GNSS-measured speed changes over a one-second window. The two distributions were harmonised onto the Targa severity boundaries, but a sensitivity analysis restricted to Targa-only events, or with a device fixed effect in the regressions, was not performed. How much of the observed spatial pattern reflects the spatial distribution of device types across the fleet therefore cannot be quantified with the current analysis.

Finally, for the cycling infrastructure case study, neighbourhood fixed effects and road-class stratification were not applied here; the reported coefficients should therefore be read as associations within this specific sample rather than as a causal ranking of design types, and any infrastructure-policy implication would require further causal investigation.

\subsection{Future directions}
Four directions for future research follow from the limitations discussed in Section~\ref{sec:limitations}. First, additional predictors -- weather, daylight, detected traffic signs, pedestrian volumes -- would improve model interpretability and capture transient effects. Second, a more advanced exposure estimate, for instance the CatBoost-based reconstruction proposed by \textcite{tomtom_historical_traffic_volumes_2025} that integrates probe counts with free-flow speeds, lane counts, functional road class, and socio-demographic attributes, would strengthen the denominator of the harsh-rate metric. Third, linking the harsh-event dataset to police-reported crashes -- not possible in this study because of data-access constraints -- would allow the surrogate to be validated against injury outcomes. Fourth, replicating the analysis in additional European cities would support a direct assessment of transferability. In addition, a formal validation of the map-matching pipeline against a held-out manually-labelled set, a sensitivity analysis to the choice of severity weights and to the device family used to record the events, and the use of multi-scale or hierarchical modelling frameworks that permit non-stationary spatial relationships are all natural follow-on steps that would sharpen several of the associations reported here into more defensible causal claims.

%% file: backmatter/credit.tex
\section*{CRediT authorship contribution statement}

\textbf{Andrea La Grotteria}: Conceptualization, Data curation, Formal analysis, Methodology, Software, Visualization, Writing – original draft, Writing – review and editing. 
\textbf{Paolo Santi}: Conceptualization, Supervision, Methodology, Writing – review and editing. 
\textbf{Titus Venverloo}: Conceptualization, Methodology, Writing – review and editing.
\textbf{Umberto Fugiglando}: Conceptualization, Funding acquisition, Writing – review and editing. 
\textbf{Carlo Ratti}: Conceptualization, Funding acquisition, Writing – review and editing.

%% file: backmatter/declarations.tex
\section*{Data availability}

The telematics records analysed in this study are proprietary to UnipolTech and were made available under a restricted research agreement; they cannot be publicly released. TomTom segment-level traffic metrics are commercially available from TomTom under licence. OpenStreetMap data are publicly available from the OpenStreetMap Foundation, and the Milan cycling-infrastructure data are available from the City of Milan open-data portal. Google Street View imagery is subject to Google's Terms of Service.

\section*{Code availability}

Analysis code is available from the corresponding author on reasonable request.

\section*{Ethics statement}

Telematics records were anonymised and aggregated at source by UnipolTech before release; no personal identifiers were accessible to the author. All analyses are performed on segment-level aggregates and contain no trip-level or vehicle-level information, consistent with GDPR requirements for the secondary use of connected-vehicle data.

\section*{Declaration of competing interest}

The author declares that he has no known competing financial interests or personal relationships that could have appeared to influence the work reported in this paper.

%% file: backmatter/acknowledgements.tex
\section*{Acknowledgements}

The authors would like to thank UnipolTech and The Urban Mobility Council for providing the data and supporting the research, TomTom for providing the data, and all members of the MIT Senseable City Consortium (including A2A, Abu Dhabi’s Department of Municipal Transportation, City of Laval, City of Rio de Janeiro, Dubai Future Foundation, FAE Technology, Hospital Albert Einstein, Seoul AI Foundation, Sondotécnica, Woven by Toyota) for supporting this research. The authors would also like to thank Prof. Patrick Thiran of EPFL for supervising the thesis work on which this article is based.

% \section*{Funding}

% This work received no specific grant from any funding agency.